\documentclass[sigconf]{acmart}
\usepackage{booktabs} % For formal tables
\usepackage{graphicx}
\usepackage{multirow}
\usepackage{amssymb,amsmath,amsfonts,scalerel}
\usepackage{amssymb}
\usepackage{color}
\usepackage{booktabs}
\usepackage{hyperref}
\usepackage[ruled,linesnumbered,vlined]{algorithm2e}
\usepackage{lineno}
\usepackage{natbib}
\usepackage{makecell}
\usepackage{arydshln}
\usepackage{xspace}
\usepackage{paralist}
\usepackage[inline]{enumitem}
\usepackage[export]{adjustbox}
\usepackage{enumitem}
\newlist{inlinelist}{enumerate*}{1}
\setlist*[inlinelist,1]{%
  label=(\roman*),
}
\author{Mohammad Aliannejadi}
\affiliation{%
  \institution{University of Amsterdam}
}
\authornote{The three authors contributed equally to this work.}
\authornote{Work done while Mohammad Aliannejadi was affiliated with Universit{\`a} della Svizzera italiana (USI).}
\email{m.aliannejadi@uva.nl}

\author{Manajit Chakraborty}
\affiliation{%
  \institution{Universit{\`a} della Svizzera italiana (USI)}
}
\authornotemark[1]
\email{manajit.chakraborty@usi.ch}

\author{Esteban Andr\'{e}s R\'{i}ssola}
\authornotemark[1]
\affiliation{%
  \institution{Universit{\`a} della Svizzera italiana (USI)}
}
\email{esteban.andres.rissola@usi.ch}

\author{Fabio Crestani}
\affiliation{%
  \institution{Universit{\`a} della Svizzera italiana (USI)}
}
\email{fabio.crestani@usi.ch}

\newcommand{\ie}{\textit{i.e., }}
\newcommand{\eg}{\textit{e.g., }}

\newcommand{\modelname}{BERT-CUR\xspace}
\newcommand{\dataname}{CAsTUR\xspace}
\newcommand{\STAB}[1]{\begin{tabular}{@{}c@{}}#1\end{tabular}}
\newcommand{\multirowrotate}[2]{\multirow{#1}{*}{\STAB{\rotatebox[origin=c]{0}{\textbf{#2}}}}}
\newcommand{\partitle}[1]{\vspace{2mm}\noindent\textbf{#1}}
\newcommand{\significant}[1]{$^{#1}$}
\newcommand{\downtri}{}
\newcommand{\uptri}{}

\begin{document}

\fancyhead{}

\title[Harnessing Evolution of Multi-Turn Conversations for Effective Answer Retrieval]{Harnessing Evolution of Multi-Turn Conversations \\ for Effective Answer Retrieval}

\begin{abstract}
With the improvements in speech recognition and voice generation technologies over the last years, a lot of companies have sought to develop conversation understanding systems that run on mobile phones or smart home devices through natural language interfaces. Conversational assistants, such as Google Assistant\texttrademark~ and Microsoft Cortana\texttrademark,~can help users to complete various types of tasks. This requires an accurate understanding of the user's information need as the conversation evolves into multiple turns. Finding relevant context in a conversation's history is challenging because of the complexity of natural language and the evolution of a user's information need. 
In this work, we present an extensive analysis of language, relevance, dependency of user utterances in a multi-turn information-seeking conversation. To this aim, we have annotated relevant utterances in the conversations released by the TREC CaST 2019 track. The annotation labels determine which of the previous utterances in a conversation can be used to improve the current one. 
Furthermore, we propose a neural utterance relevance model based on BERT fine-tuning, outperforming competitive baselines. 
We study and compare the performance of multiple retrieval models, utilizing different strategies to incorporate the user's context.
The experimental results on both classification and retrieval tasks show that our proposed approach can effectively identify and incorporate the conversation context. We show that processing the current utterance using the predicted relevant utterance leads to a 38\% relative improvement in terms of nDCG@20.
Finally, to foster research in this area, we have released the dataset of the annotations.

\end{abstract}

\copyrightyear{2020} 
\acmYear{2020} 
\setcopyright{acmcopyright}
\acmConference[CHIIR '20]{2020 Conference on Human Information Interaction and Retrieval}{March 14--18, 2020}{Vancouver, BC, Canada}
\acmBooktitle{2020 Conference on Human Information Interaction and Retrieval (CHIIR '20), March 14--18, 2020, Vancouver, BC, Canada}
\acmPrice{15.00}
\acmDOI{10.1145/3343413.3377968}
\acmISBN{978-1-4503-6892-6/20/03}

\maketitle

\section{Introduction}
Recent emergence of intelligent assistants, such as Google Assistant\texttrademark~ and Microsoft Cortana\texttrademark  ~have led to an increasing interest in research on conversational systems. Conversational assistants can help users complete various types of tasks.
The tasks can range from as simple as setting an alarm to more complex cases like health advice. Conversational assistants have been employed for information seeking~\cite{DBLP:conf/chiir/TrippasSCJS18} and recommendation~\cite{DBLP:conf/sigir/SunZ18} among other information systems applications. Moreover, such systems can be used as home assistants, \eg Alexa and Google Home; or be integrated in a smartphone \cite{DBLP:conf/sigir/AliannejadiZCC18,DBLP:conf/cikm/AliannejadiZCC18} or wearable devices, \eg Apple Siri. 
Several Information Retrieval (IR) tasks have been investigated in a conversational setting. Some examples include response ranking~\cite{DBLP:conf/sigir/YanZE17}, item recommendation~\cite{Christakopoulou:2016:TCR:2939672.2939746}, evaluation~\cite{DBLP:conf/cikm/HashemiWKZC18,DBLP:conf/www/JiangAJOZKK15}, and asking clarifying questions for users intent disambiguation~\cite{DBLP:conf/sigir/AliannejadiZCC19}. 

Although much work has been devoted towards studying single-turn conversations, several new challenges, that a multi-turn dialogue based conversational system poses, remain to be explored.
% several challenges emerge for a conversational system in a multi-turn dialogue. 
Multiple turns in a conversation can be used to understand the user information need more effectively~\cite{DBLP:conf/chiir/RadlinskiC17}. For instance, while searching for a new smartphone, user and machine could discuss various features and options in multiple turns. In order to facilitate research on multi-turn information seeking conversations, TREC introduced the Conversational Assistance Track (CAsT)\footnote{http://www.treccast.ai/} in 2019. The track contains 80 conversations on different topics, each of which consists of 8-12 turns (or utterances). 
Much work has been done on multi-turn conversational question answering~\cite{DBLP:journals/corr/abs-1808-07036,DBLP:journals/ftir/GaoGL19}. However, not much is known about how users interact with a machine in a multi-turn conversation and how their utterances can help the system to understand their information needs. 
Moreover, various aspects of multi-turn conversations are yet to be investigated, such as the way a user's information need and intent evolves as the conversation develops.

\begin{figure}[t]
    \centering
    \vspace{-0.3cm}
    \includegraphics[width=0.5\textwidth]{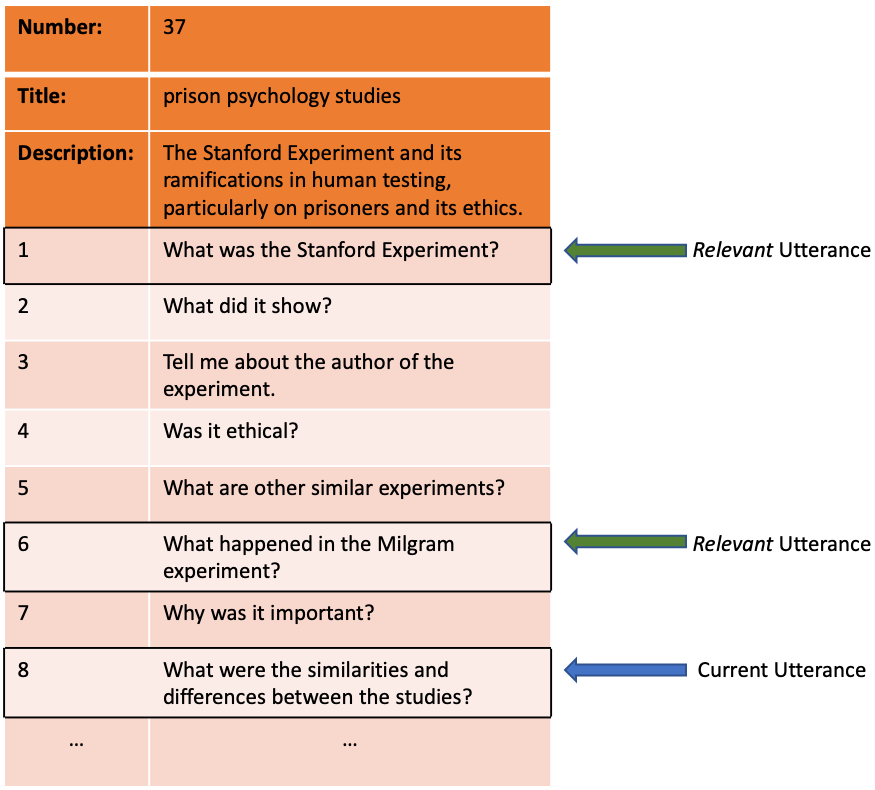}
    \vspace{-0.6cm}
    \caption{Identifying Relevant Utterances.}
    \label{fig:ques_rel}
    \vspace{-0.6cm}
\end{figure}

In this work, we aim to provide an understanding of users' behavior and how their information needs evolve during the course of a multi-turn conversation. Furthermore, we investigate the modeling of conversation context by fine-tuning BERT. To this end, we provide an in-depth analysis on the structure and evolution of users' utterances on the TREC CAsT 2019 collection. To understand the dependency of conversation turns, we have annotated each turn of the conversation with related turns in the conversation's context. This turn-relevance annotation enables us to understand and visualize the conversation evolution and dependencies. 
An utterance is \textit{relevant} to the \textit{current} one when:
\begin{inlinelist}
    \item it is needed to clarify the \textit{current} utterance or
    \item it augments the information need for the \textit{current} utterance or
    \item it entails the \textit{current} utterance.
\end{inlinelist}
As an example, consider the following conversation extracted from the TREC CAsT 2019 dataset in Figure~\ref{fig:ques_rel}. As we can observe, utterance $u_8$ cannot be answered unless additional information from previous utterances, \ie the context, is included in the question.\footnote{In this paper, we use the terms ``utterance'' and ``question'' interchangeably.} In this case, utterances $u_1$ and $u_6$\ provide the necessary context and, thus, are considered \textit{relevant} to $u_8$.  

Moreover, we propose a neural model for predicting relevant conversation turns. Our model, called \modelname, learns a joint high-dimensional representation of the user's utterance and its position in the conversation. We evaluate the effectiveness under two settings, namely, classification and retrieval performance. For the former, we study the effectiveness of our model in predicting the relatedness labels. For the latter, we follow a simple technique for incorporating the related turn into the user's current utterance. Moreover, we report the performance of classical IR models using different techniques to improve the current utterance. 

The contributions of this paper can be summarized as follows:
\begin{itemize}
    \item We create \dataname, a dataset of conversation turn relevance annotations for the TREC CAsT 2019 collection. \dataname is publicly available for research purposes.\footnote{\url{https://github.com/aliannejadi/castur}}
    \item We conduct an extensive analysis on the language, relevance, and dependency of multi-turn information-seeking conversations.
    \item We propose a neural model for predicting related conversation turns.
    \item We evaluate and analyze the performance of \modelname with two sets of experiments both as a classifier and as part of an IR system.
\end{itemize}

Experimental results show the effectiveness of our model in improving the user's utterance by adding the relevant conversation context. In particular, we see that updating utterances with their related turns results in 38\% relative improvement in terms nDCG@20. Moreover, we see in the analysis that the majority of turns are related to the first turn of the conversation. This suggests that a simple heuristic model --  assuming the first utterance of the conversation is related -- could also improve the retrieval performance. 

\section{Related Work}
Conversational search has been a long standing research problem in the IR community. However, with the recent advances in automatic voice recognition and the proliferation of Intelligent Personal Assistants such as Siri\textregistered, Alexa\texttrademark, Google Assitant\texttrademark, and Cortana\texttrademark on personal devices, the area of Conversational search has received renewed attention in the past few years. One of the earliest attempts towards conversational mode in IR can be traced back to the work of \citet{Oddy1977} who proposed the introduction of \emph{dialogue} for searching documents. \citet{Levinson1980} later proposed a system for retrieving airline information and ticket reservation using speech. Another IR system directed towards medical health professionals within a conversational setting was put forward by \citet{BOLC1985335} in 1985 for searching documents related to Gastroenterology. However, it was the work by \citet{Croft1987} on I$^3$R, an expert interface communicating with the user in a search session, which laid the foundation for conversational IR. A few years later \citet{Belkin1995} characterized information-seeking strategies for conversational IR, offering users choices in a search session based on case-based reasoning. Since then, the problem of conversational systems has been studied by researchers from both the fields of IR and Natural Language Processing (NLP) with varied interests. Conversational Agents have forayed their applications in various domains ranging from conversational recommender systems \cite{Christakopoulou:2016:TCR:2939672.2939746, DBLP:conf/sigir/SunZ18}, human memory augmentation \cite{DBLP:conf/chiir/BahrainianC18}, e-Health systems \cite{Micoulaud-Franchi2016}, personality recognition \cite{Rissola:2019} to museum tour guidance \cite{Kopp2005}. \citet{Gao2019} provides a systematic review on neural approaches to conversational AI developed in the last few years.
%Gao \etal \cite{Gao2019}
Recently, rule-based conversational IR system \cite{Loisel2012, Walker2001, williams2013dialog} have given way to learning based approaches \cite{HixonCH15} and even more recent methods based on deep learning \cite{DBLP:conf/cikm/ZhangCA0C18}. Among the several facets of conversational IR systems, one research direction is focused on analyzing user-behavior and interaction with voice-only systems \cite{Spina2017}.  Along the same line, \citet{DBLP:conf/chiir/RadlinskiC17} proposed a theoretical framework for conversational search highlighting the need for multi-turn interactions with users for narrowing down their specific information needs. \citet{DBLP:conf/chiir/TrippasSCJS18} studied conversations of real users to determine the frequently-used interactions and inform a conversational search system design. A close line of research deals with identifying user-intent while searching for information. Much work has been done in this direction, some of which include query suggestion to clarify users' intent in a traditional IR setting \cite{DBLP:conf/chiir/RadlinskiC17}, asking clarifying questions from users to understand users' intent and redirect the search \cite{DBLP:conf/sigir/AliannejadiZCC19}, clarifying user-intent by eliminating non-relevant items through negative user feedback in a conversational search \cite{DBLP:journals/corr/abs-1909-02071}. 
On the other end of the spectrum, \citet{azzopardi2018conceptualizing} posited that while understanding user intent and actions is important, little work has been addressed towards understanding the action taken by a conversational agent in the same context. They thus provided a framework for understanding the human-computer interaction from an agent's point of view. \citet{DBLP:conf/chi/VtyurinaSAC17} listed the important factors to consider while designing a conversational assistant. One of their key findings was that it is essential to maintain the conversational context which is one of the focal points of our paper. 

\section{Data Annotation and Analysis}
\subsection{TREC CAsT}
TREC CAsT was organized for the first time in 2019~\cite{DBLP:conf/trec/Dalton19}. Its main purpose is to advance research on conversational search systems. In particular, the goal of the track is to create a reusable benchmark for open-domain information centric conversational dialogues. It aspires to establish a concrete and standard collection of data with information needs to make systems directly comparable. The task was organized to focus on candidate information ranking in context involving two stages:
\begin{itemize}
    \item \textit{Read the dialogue context}: Track the evolution of the information need in the conversation, identifying salient information needed for the current turn in the conversation.
    \item \textit{Retrieve Candidate Response Information}: Perform retrieval over a large collection of paragraphs (or knowledge base content) to identify relevant information.
\end{itemize}
The track is built on top of three open-source collections, namely, MS MARCO (MAchine Reading COmprehension) Ranking passages ~\cite{DBLP:conf/nips/NguyenRSGTMD16}, TREC CAR 2018~\cite{DBLP:conf/trec/DietzG0C18} paragraph collection and News article from Washington Post (WAPO)\footnote{https://ir.nist.gov/wapo/}. MS MARCO collection was released by Microsoft in 2018 and comprises of 8,841,823 passages -- extracted from 3,563,535 web documents retrieved by Bing. TREC-CAR (Complex Answer Retrieval) is a corpus of 20 million paragraphs harvested from a snapshot of Wikipedia.

We performed experiments on MS MARCO and TREC-CAR 2018. We did not use the WAPO collection because of inconsistencies as reported by TREC and discrepancies associated with them in the relevance judgment file. The TREC CAsT 2019 track came with a set of 30 training conversations and 50 evaluation conversations\footnote{https://github.com/daltonj/treccastweb/tree/master/2019/data}. Each conversation consisted of several turns.

\subsection{Data Annotation}
\label{subsec:data_annot}
One way to determine the relevant context in a conversation is to track the evolution of the information need and identify salient information in the conversation such that the current utterance is enhanced to express user's information need more accurately. For instance, consider the conversation in Figure \ref{fig:ques_rel} and assume that the current turn is at utterance $8$. Up to this point, previous utterances, $u_1$ and $u_6$ are \textit{relevant} to $u_8$. In fact, without taking into account such information the utterance $u_8$ could hardly be addressed.
\par Given a conversation, a turn and all the previous utterances until that turn, three human annotators were asked to select one or more utterances which could help to better express the information need presented in the current turn of the conversation. After independent labeling, we computed the percentage of agreement among the annotators. If the agreement score was greater than or equal to 66.67\% (\ie at least two of the three annotators agreed on the same set of relevant utterances), we recorded the relatedness label. For turns that had an agreement score below 66.67\%, the three annotators deliberated and made an agreement on the relevance of the corresponding utterances. From the $748$ utterances, $625$ were annotated without any major disagreements. The remaining $123$ annotations were resolved by a rigorous discussion between the annotators. This procedure was conducted for both the training and evaluation sets. 

We name this set of annotated utterances as \textit{\dataname} and we released it for public use for research purposes.

\subsection{Analysis}\label{sec:DataAnalysis}
In this section, we report our analysis based on the TREC CAsT data, as well as our \dataname annotations. 

\partitle{Basic statistics.}
In the first half of Table \ref{tab:basicstats} we present a summary statistics of TREC CAsT 2019 conversations dataset. We note that on average conversations span over nine turns with an average length of seven terms per turn. In particular, we notice that there are only 19 conversations out of 80, which have eleven turns or more. In the second half of the table, we outline statistics associated with \dataname (Section \ref{subsec:data_annot}). We observe that $18\%$ of the utterances do not have any related utterances. This indicates that those utterances are either self-contained or no additional information in previous conversation turns existed. Majority of the utterances have only one relevant turn. Very few of the utterances ($< 1\%$) have three or more relevant turns.

\begin{table}[t]
    \centering
    \caption{Descriptive Statistics for TREC-CAst and \dataname.}
    \vspace{-0.6cm}
    \begin{tabular}{llr}\\ 
        \toprule
        \multicolumn{2}{l}{Parameter} & Value \\ 
        \midrule
        \multicolumn{2}{l}{\# conversations}  & 80 \\
        \multicolumn{2}{l}{\# turns} & 748 \\
        \multicolumn{2}{l}{Avg. \# turns per conversation} & 9.35 $\pm$ 1.32 \\
        \multicolumn{2}{l}{Median \# turns per conversation} & 9 \\
        \multicolumn{2}{l}{Avg. \# terms per turn} & 7.21 $\pm$ 2.02\\
        \hdashline
        \multicolumn{2}{l}{\# relevance annotations} & 752 \\
        \multicolumn{2}{l}{\% utterances with \textbf{0} rel. turns} & 18\% \\
        \multicolumn{2}{l}{\% utterances with \textbf{1} rel. turns} & 71\% \\
        \multicolumn{2}{l}{\% utterances with \textbf{2} rel. turns} & 10\% \\
        \multicolumn{2}{l}{\% utterances with \textbf{3} rel. turns} & $<$ 1\% \\
        \multicolumn{2}{l}{\% utterances with \textbf{4} rel. turns} & $<$ 1\% \\
        \bottomrule
    \end{tabular}
    \label{tab:basicstats}
    \vspace{-0.5cm}
\end{table}

\begin{figure}[t]
    \centering
    \vspace{-0.4cm}
    \includegraphics[width=\columnwidth]{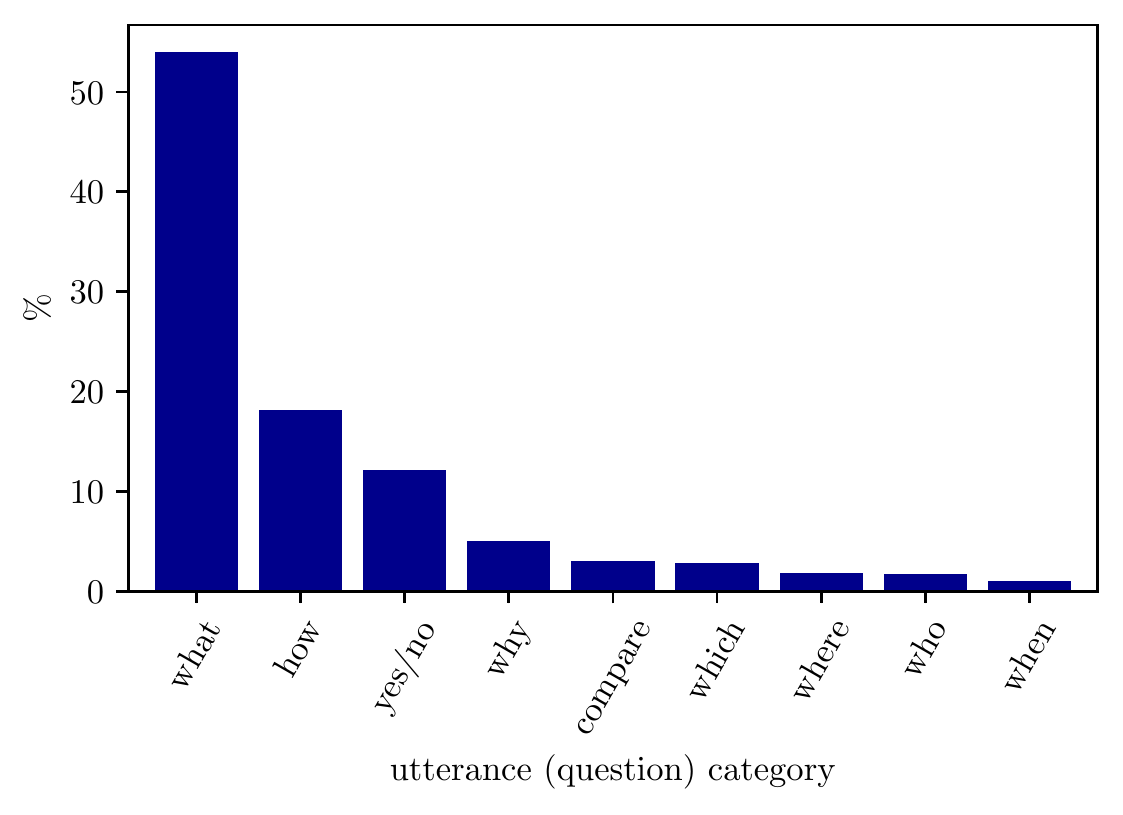}
    \vspace{-0.8cm}
    \caption{Question category distribution.}
    \label{fig:dist_qcat}
    \vspace{-0.3cm}
\end{figure}

\partitle{Question language and categories.}
Depending on the type of question, we manually segregate them into one of the following nine categories: \textit{What}, \textit{When}, \textit{Where}, \textit{Which}, \textit{Who}, \textit{Why}, \textit{How}, \textit{Yes/No} and \textit{Compare}. The category \textit{What} includes all questions that either starts\footnote{The only exception are comparison questions like `What are the similarities/differences [...] '} with a ``What'' or ask for a description like ``Tell me about Boer goats.'' Questions that either collate or contrast two or more entities are categorized into \textit{Compare}, \eg ``What were the similarities and differences between the studies?'' The rest of the categories are self-explanatory. The distribution of questions across categories is presented in Figure \ref{fig:dist_qcat}. Additionally, we computed for each category the number of conversations in which it is the leading category. We found that \textit{What} questions dominate 67 conversations, while \textit{How} dominates 12 conversations and the remaining one is dominated by \textit{Why} questions.

\begin{figure}[!ht]
    \centering
    \vspace{-0.4cm}
    \includegraphics[width=1.1\columnwidth]{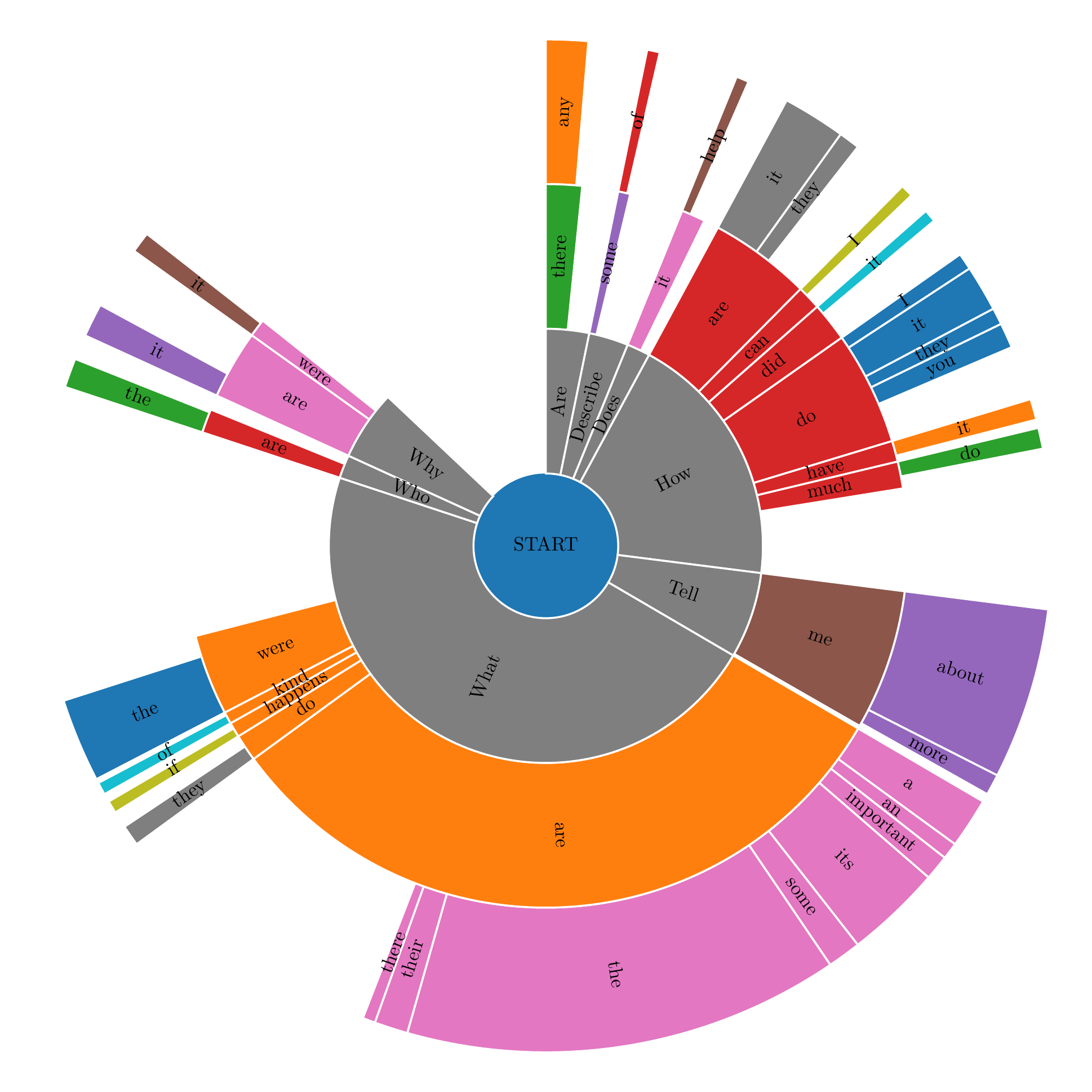}
    \vspace{-0.6cm}
    \caption{Distribution of top 40 trigrams of the questions (best viewed in color).}
    \label{fig:question_sunburst}
    \vspace{-0.5cm}
\end{figure}

In Figure \ref{fig:question_sunburst}, we present the distribution of top 40 trigrams of the questions across all conversations. This diagram helps us identify the share and sequence of terms in a question across all conversations. To provide a more meaningful representation some of the auxiliary verbs have been modified (\eg ``is'' $\rightarrow$ ``are'').
For example, we see that many of the utterances start with ``What'' and we see that in many cases the users are asking for more details or additional information about specific topics (\eg ``What are its'' and ``What happens if'').
On closer observation we find that the third word in the trigrams are usually pronouns, which indicates that they are referring to the context of the conversation. 
In fact, we found that 50\% of CAsT utterances contained at least one kind of English pronouns, 42\% of which was the third word in the sentence.
This highlights the significance of determining \textit{relevant} context in a conversation.

% \popline

\begin{figure}[t]
    \centering
    \vspace{-0.4cm}
    \includegraphics[width=\columnwidth]{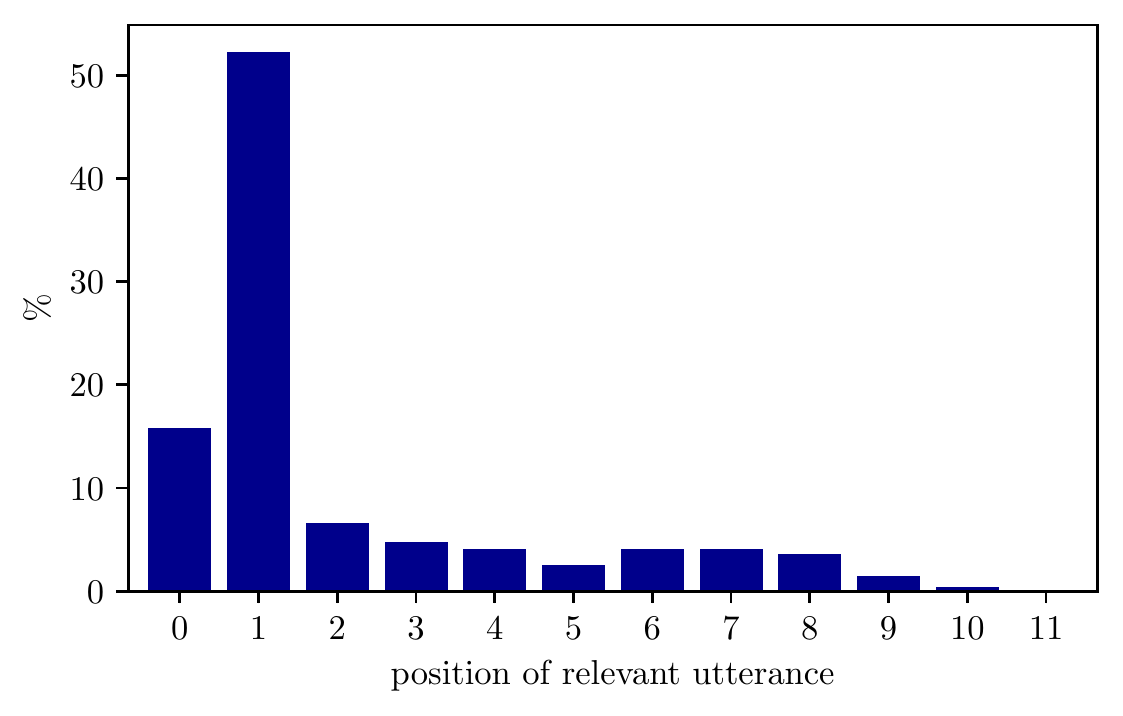}
    \vspace{-0.8cm}
    \caption{Distribution of \textit{relevant} turn number $\mathcal{P}_{RT}$~.}
    \label{fig:relq}
    \vspace{-0.3cm}
\end{figure}

\partitle{Distribution of relevant turns.}
We also analyze the position of relevant turn, $\mathcal{P}_{RT}$, for each of the turns and how they are distributed across all conversations (Figure \ref{fig:relq}). We find that for more than 50\% of the total turns, a relevant turn is found at the first position in the conversation. 

The position 0 in the figure represents the number of self-contained utterances, \ie the ones which do not have any prior relevant utterance. This position does not account for the first utterances in each conversation, since they are assumed to have no \textit{relevant} priors. Hence, they constitute 16\% of the total number of relevant utterances.

\begin{figure}[!ht]
    \vspace{-4mm}
    \includegraphics[width=\columnwidth]{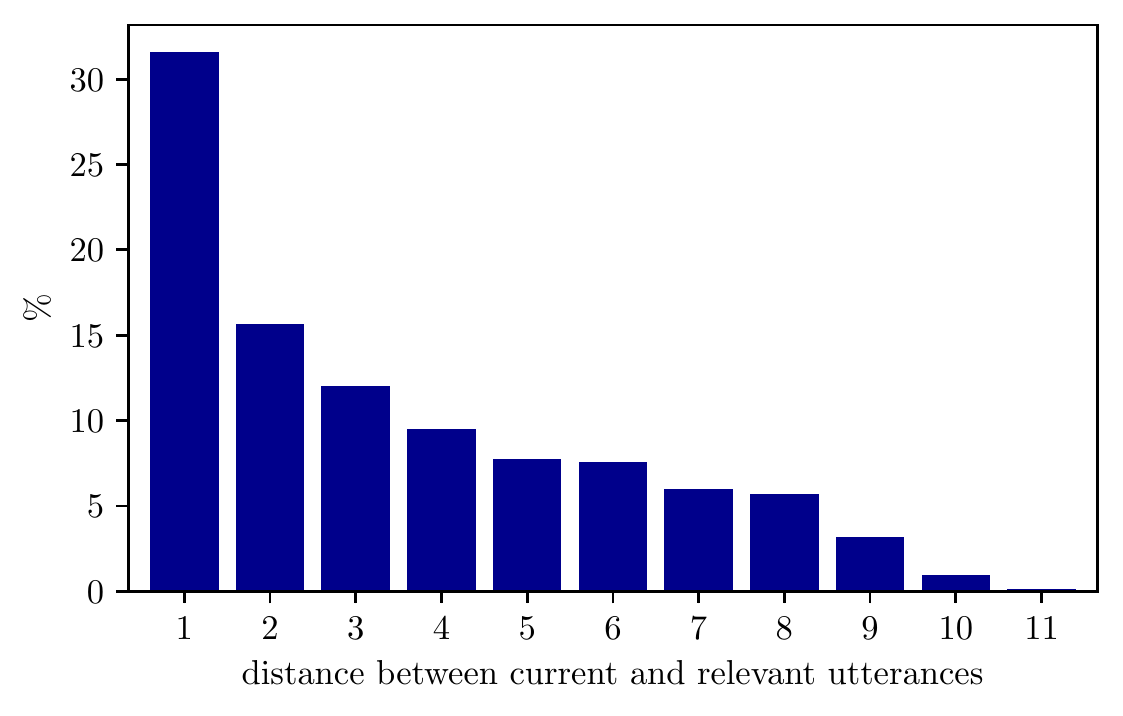}
    \vspace{-0.6cm}
    \caption{Distribution of difference in position of current and relevant question $\Delta\mathcal{P}_{(CT,RT)}$~.}
    \label{fig:distrelq}
    \vspace{-0.5cm}
\end{figure}

\partitle{Evolution of conversation.}
Since the topic of the conversation generally evolves or changes with each turn, we want to observe the difference in position of the relevant turn ($RT$) with the current turn ($CT$), $\Delta\mathcal{P}_{(CT,RT)}$, where $\mathcal{P}$ denotes position of an utterance. We find that for a large share of utterances, a relevant turn is found in close proximity (Figure \ref{fig:distrelq}). In fact, as the figure suggests, more than 30\% of the utterances have a relevant turn at a position immediately prior to them.

\begin{figure}[!ht]
    \centering
    % \vspace{-0.6cm}
    \includegraphics[width=\columnwidth]{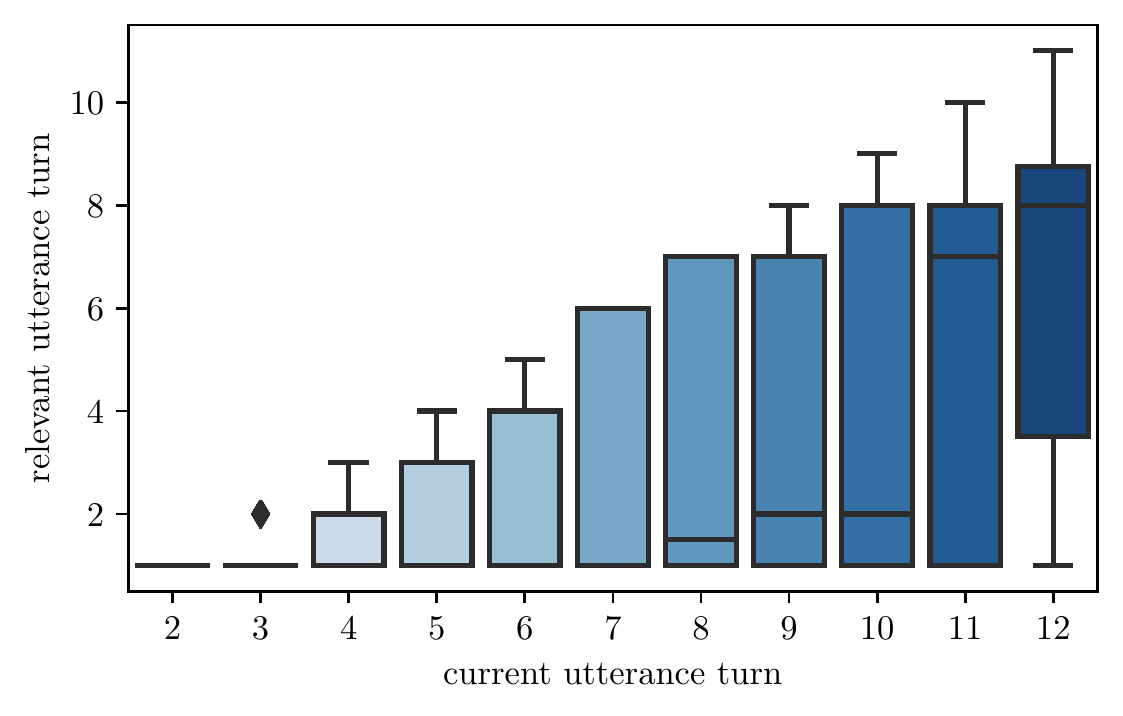}
    \vspace{-0.6cm}
    \caption{Distribution of relevant turn ($\mathcal{P}_{RT}$) per current turn ($\mathcal{P}_{CT}$).}
    \vspace{-0.55cm}
    \label{fig:box_relq}
\end{figure}

Next, we study the relation between the position of the current utterance and its relevant turn. For instance, as observed in Figure \ref{fig:box_relq}, for turn number 10, the average relevant question position is 2; while for turn 11, the relevant question is usually found at position 7 in the conversation. 
This suggests that in most conversations of the TREC CAsT, the first seven turns are mostly dependent on the first turn, suggesting that users explore a single topic in several turns. However, we can also see that as the conversation progresses, the later turns are less dependent on the first turn (\eg turns 10, 11, and 12). This indicates the possibility that longer conversations may include multiple topics (\ie either elaborating a subtopic or starting a new topic). 

\begin{figure}[t]
    \centering
    \vspace{-4mm}
    \includegraphics[width=\columnwidth]{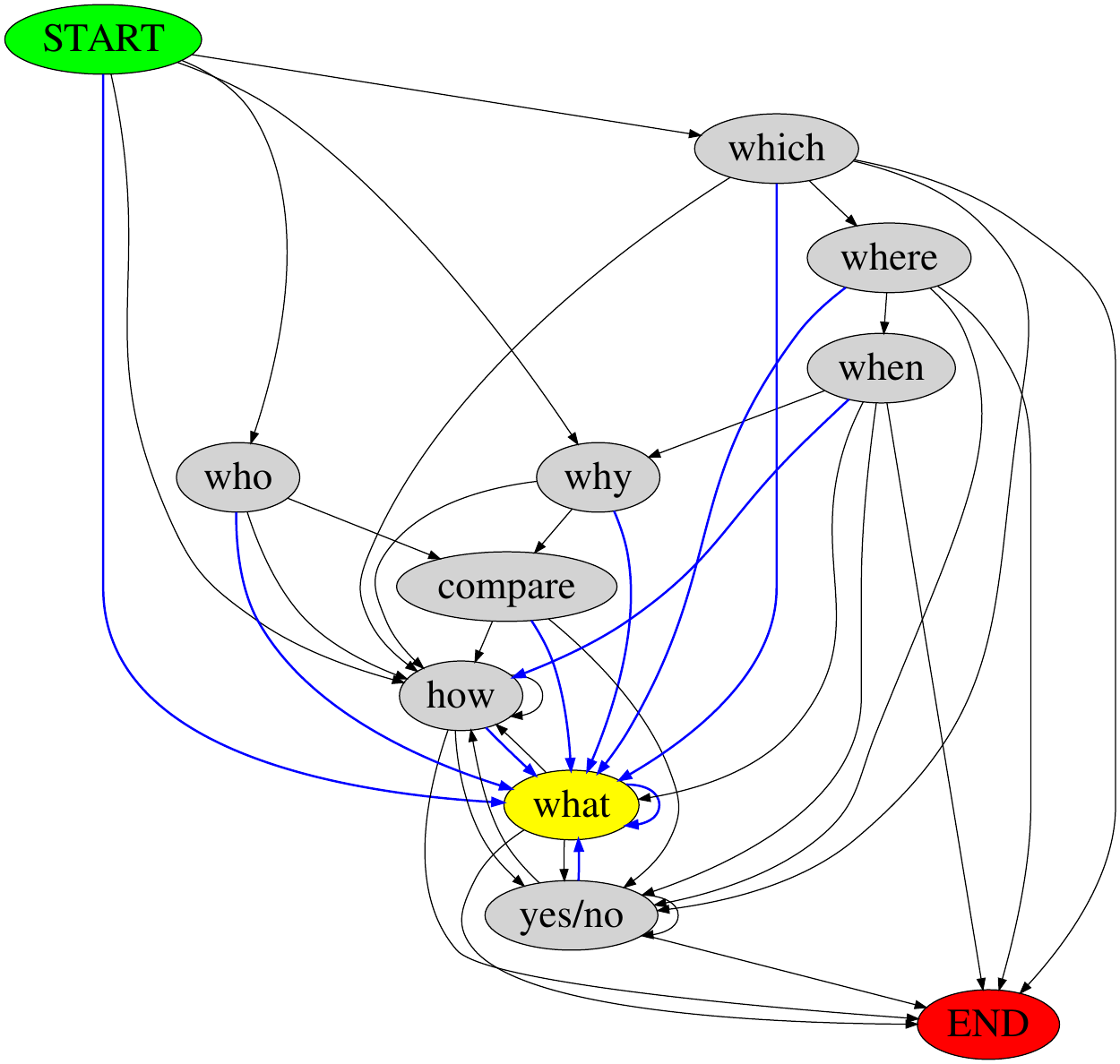}
    \vspace{-0.6cm}
    \caption{Flow pattern of conversations with a Markov model. Nodes are question categories. Edges are directed and weighted by transition probability (best viewed in color). }
    \vspace{-0.4cm}
    \label{fig:markov_conv}
\end{figure}

\partitle{Dependency among question categories.}
We use a Markov Model to analyze the flow patterns in the conversations
as shown in Figure~\ref{fig:markov_conv}. The aim of this analysis is to understand the natural progression of a conversation in terms of question categories. Each of the nine question categories forms a node in a graph with two additional placeholder nodes, START and END which denote the beginning and end of the conversation. Therefore, for the conversation example in Figure \ref{fig:ques_rel}, the labeled sequence would be ``START$\rightarrow$What$\rightarrow$What$\rightarrow$What$\rightarrow$Yes/No$\rightarrow$\\What$\rightarrow$What$\rightarrow$Why$\rightarrow$Compare$\rightarrow$ ... $\rightarrow$END''. The edges between the nodes are weighted by the transition rate. 
We denote the highest probable transition from each node by marking its respective edge in blue (\eg ``which $\rightarrow$ what'').
This figure represents the question category transition pattern in an information seeking process. We can make some observations from
the graph: \begin{inparaenum}
\item For most cases, we note that the conversation starts with a \textit{What} question while there is no clear candidate for end of conversation.
\item \textit{What} questions are more likely to be followed by another \textit{What} question, indicated by a blue self-loop.
\item \textit{Which} questions are followed by more diversified questions \ie they can be followed by different category questions.
\end{inparaenum}

\section{Retrieval Framework}

In this section, we describe the framework that we design for utterance dependency prediction and document retrieval. First, we explain our model for predicting the related utterance to the current one. Further, we explain how we use the prediction to improve the current utterance and retrieve documents.

\subsection{Utterance Relevance Prediction Model}\label{sec:utter_rel}

We now describe our BERT\footnote{BERT: Bidirectional Encoder Representations from Transformers~\cite{DBLP:conf/naacl/DevlinCLT19}} Utterance Relevance Prediction model, called \modelname. Here, our goal is to maximize the performance of our retrieval model by selecting the most relevant utterances as the context of the current utterance. The aim is to select related utterances such that the current utterance is enhanced to express the user information need more accurately. 
As we saw in Section~\ref{sec:DataAnalysis}, predicting the relevant context is critical as many of the utterances contain a pronoun, referring to their context in the conversation. Based on our analysis, the relative distance of two utterances is negatively correlated with their relevance. Therefore, finding the related utterance among multiple ones in the conversation history requires joint modeling of the utterance language and position in the conversation.  
We learn a high-dimensional language and position representation for predicting the most relevant utterances to the current utterance. Formally, \modelname estimates the probability $p(R = 1|u_i,u_j)$, where $R$ is a binary random variable indicating whether the utterance $u$ should be marked as relevant ($R=1$) or not ($R=0$). Let $u_i$ and $u_j$ denote the current and the candidate utterance, respectively. The utterance relevance probability in the \modelname model is estimated as follows:
\vspace{-0.15cm}
\begin{equation}
    p(R = 1|u_i,u_j; i, j) = \psi\big(\phi_{U}(u_i), \phi_{C}(u_j), \eta_{U}(i), \eta_{C}(j)\big)~,
    \vspace{-0.1cm}
\end{equation}
\noindent where $\phi_{U}$ and $\phi_{C}$ respectively denote the current and candidate utterance representations. Whereas $\eta_{U}$ and $\eta_{C}$ denote the position representation of the current and candidate utterances, respectively. $\psi$ is the matching component that takes the aforementioned representations and produces an utterance relevance score. There are various ways to implement any of these components. 

We implement $\phi_{U}$ and $\phi_{C}$ similarly using a function that maps a sequence of words to a $d$-dimensional representation ($V^s \rightarrow \mathbb{R}^d$). We use the BERT~\cite{DBLP:conf/naacl/DevlinCLT19} model to learn these representation functions. BERT is a deep neural network with 12 layers that uses an attention-based network called Transformers~\cite{DBLP:conf/nips/VaswaniSPUJGKP17}. We initialize the BERT parameters with the model that is pre-trained for the language modeling task on Wikipedia and fine-tune the parameters on \dataname. BERT has recently outperformed state-of-the-art models in a number of language understanding and retrieval tasks~\cite{DBLP:conf/naacl/DevlinCLT19}.
In addition, BERT shows promising results in modeling short texts and small collections.

We model the utterance position using $\eta_{U}$ and $\eta_{C}$, mapping the conversation turn position to a high-dimensional representation ($\mathbb{N}^t \rightarrow \mathbb{R}^{dt}$).
The component $\psi$ is modeled using a fully-connected feed-forward network with the output dimensionality of 2. Rectified linear unit (ReLU) is employed as the activation function in the hidden layers, and a softmax function is applied on the output layer to compute the probability of each label (\ie relevant or non-relevant). To train \modelname, we use a cross-entropy loss function and follow the point-wise learning approach. We train \modelname using two BERT models. \modelname$_{\text{BASE}}$ fine-tunes the uncased base BERT model, whereas \modelname$_{\text{LARGE}}$ fine-tunes the uncased large BERT model.
\subsection{Utterance Reformulation and Document Retrieval}

After finding the relevant utterances as conversation context, the next step is
to augment and reformulate the current question for improved document retrieval. This step involves processing of the original utterance, as well as adding terms from its context to help the document retrieval model rank relevant documents higher. The query processing and reformation is done in the following stages:

\begin{enumerate}
    \item In a conversation, it is imperative that there will be some form of relationship between the turns. We observed from the data that a large share of utterances were dependent on previous turns in the conversation. For example, in Figure \ref{fig:ques_rel}, $u_2$ is clearly dependent on $u_1$; $u_3$, $u_4$ also depends on $u_1$ to make a meaningful question. Hence, it is essential to perform anaphora (coreference) resolution on such utterances to make them a complete question in itself. This step is performed only on FirstUtterance (FU) technique. To this end, we use AllenNLP~\cite{Gardner2017AllenNLP} coreference resolution tool . For self-contained and first turns in the conversations, the output of the tool is the same as the original turn. 
    \item On this modified query, we perform stopword removal, removal of special characters, and tokenization.
    \item We perform an additional step on utterances in \modelname to ensure that the utterances retain the same order in the final query as in the conversation. This procedure ensures that BERT learns from the relative ordering of questions as well. It is to be noted that for our proposed \modelname model, we did not perform Step 1 on the utterances since our objective is to learn the correct relevant utterance from the context.
\end{enumerate}

After updating the utterances, we pass the resulting query set to various document retrieval models. For document retrieval we employ Galago\footnote{https://www.lemurproject.org/galago.php} and index both the MS MARCO and TREC-CAR collections together. The total number of documents indexed was 38,429,852. 
We employ four different techniques to reformulate the current utterance and compare the performance of three term-matching IR models, namely, Okapi-BM25~\cite{DBLP:conf/trec/RobertsonWJHG94}, Query Likelihood (QL)~\cite{DBLP:conf/sigir/PonteC98} and Divergence From Randomness (DFR)~\cite{DBLP:journals/tois/AmatiR02}. We investigate four utterance selection strategies for each of the retrieval models. More details of the utterance reformulation strategies can be found in Table~\ref{tab:ureform}. As we see in Table~\ref{tab:ret_res}, the performance of the proposed \modelname is compared with the original utterance (\ie \textit{Orig.}), as well as the reformulated utterance using the first utterance of the conversation (\ie \textit{FU}). We use the first utterance as a heuristic model, motivated by our analysis which showed that the majority of relevant utterances are found in the first turn of the conversation. Comparison with oracle models allows us to validate the effectiveness of utterance relevance annotations. Also, it gives us an idea of the upper bound in improving the retrieval performance, following the mentioned utterance reformulation steps.

Moreover, it is worth mentioning that we do not investigate the performance of neural ranking or re-ranking approaches such as BERT fine-tuning models. Our decision is motivated by the relevance annotation used to create the TREC CAsT collection. Since the pooling and labeling is done using only term-matching models, the superiority of more sophisticated approaches such BERT is not necessarily reflected in the evaluation, simply because the documents that they retrieve were not present in the evaluation pool. Therefore, the validity of the experiments can be guaranteed only on term-matching approaches. 

\begin{table}[t]
    \centering
    % \vspace{-0.4cm}
    \caption{Names and descriptions of utterance reformulation techniques.}
    \vspace{-0.3cm}
    \begin{tabular}{ll}
         \toprule
         Name & Description  \\
         \midrule
         \makecell[t]{\textbf{Orig.}} & The original unmodified utterance. \\
         \makecell[t]{\textbf{FU}}   &  \makecell[l]{The current utterance reformulated using the first \\ utterance of the conversation.} \\
         \makecell[t]{\textbf{PU}}   &  \makecell[l]{The current utterance reformulated using the immediate\\ previous utterance of the conversation.} \\
         \makecell[t]{\textbf{FPU}}   &  \makecell[l]{The current utterance reformulated using the first and\\ immediate previous utterance of the conversation.} \\
         \makecell[t]{\textbf{AU}}   &  \makecell[l]{The current utterance reformulated using all the previous\\ utterances of the conversation until that point.} \\
         \makecell[t]{\textbf{PrU}}  & \makecell[l]{The current utterance reformulated using the relevant \\ utterance predicted by \modelname.} \\
         \makecell[t]{\textbf{Oracle}} & \makecell[l]{The current utterance reformulated using \\ the true relevance labels.} \\
         \bottomrule
    \end{tabular}
    \vspace{-0.5cm}
    \label{tab:ureform}
\end{table}

\section{Experiments}
In this section, we describe the data, metrics, and compared methods. Furthermore, we evaluate the performance of the proposed models in comparison with the baselines. We also study the performance of the models at different turns of the conversation.

%%%%%%%%%%%%%%%%%%%%%%%%%%%%%%%%%%%%%%%%%%% from other papers
\subsection{Experimental Setup}

\partitle{Dataset.} 
Since we conduct the evaluation on the TREC CAsT data, we did not have the relevance judgment for the 50 evaluation topics provided by TREC, at the point of writing this paper. In the absence of gold standard, we used the 30 training topics (which had 269 utterances in total) from the TREC challenge as our \textit{test} set while using the TREC evaluation topics as our \textit{training} set for \modelname. Out of 50 training topics, which had 479 utterances, we randomly selected 10\% as \emph{dev} set. We used these sets to train, tune, and test \modelname, as well as the compared methods.

\partitle{Utterance relevance prediction evaluation metrics.} 
Since we model the utterance relevance problem as classification, we consider three standard metrics to evaluate the effectiveness of the models, namely, Precision (Prec.), Recall and F1-Measure. For computing these scores, we treat each of the 748 turns (\ie as a single instance) individually.                                

\partitle{Ad-hoc retrieval evaluation metrics.} 
Here, effectiveness is measured considering the standard evaluation metrics of IR, as well as the ones used in the TREC CAsT track.
Therefore, we evaluate the retrieval models using eight standard evaluation metrics: Mean Average Precision (MAP), Mean Reciprocal Rank (MRR), normalized discounted cumulative gain at the top 5,10 and 20 retrieved documents (nDCG@5, nDCG@10, nDCG@20) and, precision at the top 5,10 and 20 retrieved documents (P@5, P@10 and P@20).
We use the relevance assessments as released by TREC. 

\partitle{Statistical test.} We determine statistically significant differences using the two-tailed paired t-test at a $95\%$ confidence interval ($p < 0.05$).

\partitle{Compared methods.} 
We compare the performance of \modelname with competitive classification baseline methods, details of which are as follows:
\begin{itemize}[leftmargin=*]
\item[--] \textit{FirstUtterance/FirstPrevUtterance/AllPrevUtterance}: we always select the first/first and previous/all previous  utterances in the conversation as relevant to the current one.
\item[--] \textit{Random}: we select any random turn from the conversation until the current utterance to be \textit{relevant}.
\item[--] \textit{k-NN, k-NN-AWE}: To find the nearest neighbors in \textit{k}-nearest neighbors (k-NN), we consider the cosine similarity between the TF-IDF vectors and turn numbers of utterances. Then, we take the labels of the nearest questions and produce the relevant question ranking. As for k-NN-AWE, we compute the cosine similarity between the average word embedding (AWE) of the questions obtained from GloVe~\cite{DBLP:conf/emnlp/PenningtonSM14} with 300 dimensions. The value of \textit{k} is determined using 5-fold cross validation on the \textit{dev} set.
\item[--] \textit{MLP, MLP-AWE}: For MLP we used 2 hidden layers along with Adam optimization algorithm with default parameters. The MLP-AWE method for using word embeddings was similar to the one used in k-NN-AWE. 
\item[--] \textit{SVM, SVM-AWE}: We use linear SVM with $C=0.1$ to predict the most relevant question. Similar to MLP-AWE, SVM-AWE also follows the same approach as in k-NN-AWE for obtaining the average word embeddings.
\end{itemize}

\vspace{-0.2cm}
\subsection{Results and Discussion}
We evaluate the performance of \modelname and the utterance reformulation techniques with respect to classification and document retrieval performance. Furthermore, we conduct various analyses on the results to understand the influence of conversation context in the performance.

\partitle{Relevant utterance prediction performance comparison.} We start by presenting the comparison of the relevant utterance prediction models in Table \ref{tab:classficiation}. We can clearly observe that our \modelname outperforms all the other prediction models. Although, the \textit{SVM-AWE} model achieves the same Precision as \modelname$_{\text{BASE}}$, our proposed model supersedes the \textit{SVM-AWE} in terms of Recall and hence F1-Measure as well. In terms of F1-Measure, the closest performing model to \modelname is \textit{FirstUtterance}, which corroborates our findings from Figure \ref{fig:relq} that first utterance is relevant for more than 50\% of the total utterances in the dataset. It states that in the majority of cases, the first turn in the conversation identifies the main topic of the conversation and the rest of the turns explore more details of the same topic, rather than shifting to a different topic. Notice that \textit{FirstUtterance} is also the best performing heuristic for utterance prediction. 
Also, we observe that \modelname$_{\text{LARGE}}$ also outperforms the baseline models. However, unlike other tasks, we observe better performance achieved by fine-tuning the BERT base model. This is because of smaller data size compared to the tasks where the BERT large model performs better.

\begin{table}[t]
    \centering
    
    \caption{Question Relevance Prediction Performance. Best performances are marked in bold.}
    \begin{tabular}{lrrr}
        \toprule
        & \multicolumn{1}{c}{Prec.} & \multicolumn{1}{c}{Recall} & \multicolumn{1}{c}{F1} \\
        \midrule
        FirstUtterance & 0.60 & 0.65 & 0.62 \\
        PrevUtterance & 0.29 & 0.31 & 0.30 \\
        FirstPrevUtterance & 0.42 & 0.85 & 0.56 \\
        AllPrevUtterances & 0.20 & \textbf{1.00} & 0.33 \\
        Random & 0.20\downtri & 0.50\downtri & 0.29\downtri \\
        % \textbf{PreviousQuestion} \\
        
        k-NN &  0.64\uptri & 0.49\downtri & 0.55\downtri\\
        MLP & 0.63\uptri & 0.48\downtri & 0.54\downtri\\
        SVM & 0.64\uptri & 0.43\downtri & 0.52\downtri\\
        
        k-NN-AWE & 0.64\uptri & 0.49\downtri & 0.55\downtri \\
        MLP-AWE & 0.64\uptri & 0.53\downtri & 0.58\downtri \\
        SVM-AWE & \textbf{0.66}\uptri & 0.49\downtri & 0.57\downtri\\
        
        \hdashline
        \modelname$_\text{BASE}$ & \textbf{0.66\uptri} & 0.69\uptri & \textbf{0.67\uptri}\\
        \modelname$_\text{LARGE}$ & 0.65\uptri & 0.67\uptri & 0.66\uptri \\
        \bottomrule
    \end{tabular}
    \vspace{-0.4cm}
    \label{tab:classficiation}
\end{table}

\partitle{Document retrieval performance comparison.} Here, we also study the performance of an oracle model, \ie assuming that an oracle model is aware of the correct relevant question to all the questions. The goal is to show to what extent adding a relevant question to the current turn can improve the performance of a retrieval system. The results are presented in Table \ref{tab:ret_res}. Seven utterance reformulation techniques (see Table~\ref{tab:ureform}) are compared for each of the three retrieval models. 
The best retrieval performance is achieved when we use the QL model for retrieval.
We see that all utterance reformulation strategies improve the performance compared to the original utterance, indicating the high dependence of utterances on previous turns of the conversation. Also, we see that even employing simple heuristic utterance selection, such as selecting the first utterance (\ie FU) or previous utterance (\ie PU) of the conversation, leads to retrieval improvement. However, it is interesting to note that no heuristic method achieves significant improvement over the original utterance model (\ie Orgi.).

As we see in the sixth row of each section of Table \ref{tab:ret_res}, our proposed model \textit{PrU}, outperforms all the baselines for all the three retrieval models.
We observe that while the improvements on the DFR and QL models are consistently statistically significant (except for MRR on QL), they are less consistent on BM25. This could be due to the length of the reformulated utterances which are usually augmented with the relevant utterance(s). Notice in the table that the improvements over \textit{FU} is statistically significant when we use the QL retrieval model, in terms of P@$k$. 
This shows the potential gain of adding good relevant utterances to the current utterance on the performance of a conversational system. It is worth mentioning that although Oracle model is an idealistic model, there were three instances in which \textit{PrU} performed better than \textit{Oracle} model \ie for DFR on P@10, as well as for QL on nDCG@5 and P@5. This suggests the strength of our model in capturing the semantic dependency of utterances, achieving comparable performance to the human annotators. It is worth noting that we report the performance of our proposed \modelname model for precision-oriented relevant utterance selection (explained in details later in this section). 
Interestingly, we see that the difference between \textit{Oracle} and \textit{PrU} are not significant (no $^6$ in Table~\ref{tab:ret_res}). This again suggests that \modelname can effectively model and capture the semantic dependencies of utterances in different turns to the point where it can help the retrieval model to achieve comparable results to the perfect human annotation.

\begin{table*}[t]
    \centering
    % \vspace{-0.4cm}
    \caption{Passage Retrieval Performance Comparison. The superscripts 1/2/3/4/5/6 denote that the
improvements over Orig./FU/PU/FPU/AU/PrU are statistically significant ($p < 0.05$). Best performing runs except Oracle model are marked in bold.}
\vspace{-0.4cm}
    \begin{tabular}{llllllllll}
        \toprule
        \multicolumn{2}{c}{Model}  & \multicolumn{1}{c}{MAP} & \multicolumn{1}{c}{MRR} & \multicolumn{1}{c}{nDCG@5} & \multicolumn{1}{c}{nDCG@10} & \multicolumn{1}{c}{nDCG@20} & \multicolumn{1}{c}{P@5} & \multicolumn{1}{c}{P@10} & \multicolumn{1}{c}{P@20} \\
        \cmidrule{1-10}
        \multirowrotate{7}{DFR}
        & Orig. &  0.1079 & 0.1497 &  0.0988 &  0.1086 &  0.1319 &  0.0950 &  0.0800 &  0.0621 \\
        & FU & 0.1682\significant{5} &  0.2480 &  0.1687\significant{5} &  0.1840\significant{5} &  0.2189\significant{5} &  0.1450\significant{45} &  0.1108\significant{45} &  0.0858\significant{5} \\
        & PU & 0.1438 & 0.2072 & 0.1399  & 0.1505  & 0.1793  &  0.1233 & 0.0983 & 0.0762\\
        & FPU & 0.1487 &  0.2312 &  0.1387 & 0.1654  & 0.1963  & 0.1050 &  0.0942 & 0.0763   \\
        & AU & 0.1226  & 0.1923  & 0.1201  &  0.1299 & 0.1573  & 0.1017 & 0.0842  & 0.0667\\
        
        & PrU & \textbf{0.1919\significant{1345}} &  \textbf{0.2754\significant{135}} &  \textbf{0.1906\significant{1345}} &  \textbf{0.2058\significant{135}} &  \textbf{0.2418\significant{135}} &  \textbf{0.1650\significant{1345}} &  \textbf{0.1317\significant{1345}} &  \textbf{0.1004\significant{1345}} \\
        
        & Oracle & 0.2009\significant{12345} &  0.3004\significant{12345} &  0.2004\significant{1345} &  0.2154\significant{1345} &  0.2551\significant{12345} &  0.1717\significant{1345} &  0.1308\significant{12345} &  0.1038\significant{12345} \\
          
        \midrule
        \multirowrotate{7}{BM25}
        & Orig. & 0.0887 & 0.1420 & 0.0841 & 0.0948 & 0.1131 & 0.0817 & 0.0733 & 0.0554\\
        & FU & 0.1359\significant{5} &  0.2287\significant{5} &  0.1432\significant{5} &  0.1668\significant{5} &  0.1907\significant{5} &  0.1250\significant{45} &  0.1092\significant{45} &  0.0788\significant{5} \\
        & PU & 0.1263 & 0.1962  & 0.1268  & 0.1364  & 0.1679  & 0.1083  & 0.0908 & 0.0725  \\
        & FPU & 0.1272 & 0.2108  & 0.1301  & 0.1472  & 0.1754  & 0.0967  & 0.0883 & 0.0692 \\
        & AU & 0.1014 & 0.1654  & 0.1013  & 0.1135  & 0.1374  & 0.0883  & 0.0783 & 0.0629 \\
        & PrU & \textbf{0.1546}\significant{5} &  \textbf{0.2465}\significant{5} &  \textbf{0.1548}\significant{5} &  \textbf{0.1805\significant{135}} &  \textbf{0.2097}\significant{5} &  \textbf{0.1400}\significant{45} &  \textbf{0.1258\significant{1345}} &  \textbf{0.0921\significant{1345}} \\
        
        & Oracle & 0.1633\significant{1345} &  0.2658\significant{345} &  0.1671\significant{1345} &  0.1925\significant{1345} &  0.2225\significant{1345} &  0.1450\significant{1345} &  0.1267\significant{1345} &  0.0933\significant{12345}\\
        
        \midrule
        \multirowrotate{7}{QL} 
        & Orig. & 0.1290 & 0.1603 & 0.1115 & 0.1302 & 0.1681 & 0.1100 & 0.0933 & 0.0833\\
        & FU & 0.2026\significant{5} &  0.2788 &  0.1726\significant{5} &  0.2111\significant{5} &  0.2602\significant{5} &  0.1500\significant{5} &  0.1408\significant{5} &  0.1113\significant{5} \\
        & PU & 0.1745 & 0.2486  & 0.1522  & 0.1799  & 0.2164  & 0.1433  & 0.1300 &  0.0963 \\
        & FPU & 0.1884\significant{5} & 0.2735\significant{5}  & 0.1602\significant{5}  & 0.1968\significant{5}  & 0.2368\significant{5}  & 0.1250  & 0.1275\significant{5} &  0.0979\significant{5} \\
        & AU & 0.1429 & 0.2220  & 0.1211  & 0.1519  & 0.1848  & 0.1083  & 0.1083 & 0.0854  \\
        
        & PrU & \textbf{0.2362\significant{1345}} &  \textbf{0.3009}\significant{5} &  \textbf{0.2082\significant{1345}} &  \textbf{0.2470\significant{1345}} &  \textbf{0.2984\significant{1345}} &  \textbf{0.1883\significant{12345}} &  \textbf{0.1658\significant{12345}} &  \textbf{0.1321\significant{12345}} \\
        
        & Oracle & 0.2435\significant{12345} &  0.3243\significant{12345} &  0.2059\significant{1345} &  0.2493\significant{12345} &  0.3103\significant{12345} &  0.1817\significant{12345} &  0.1692\significant{12345} &  0.1383\significant{12345}\\
        \bottomrule
    \end{tabular}
    \label{tab:ret_res}
    \vspace{-1mm}
\end{table*}

\begin{figure}[t]
    \centering
    \includegraphics[width=\columnwidth]{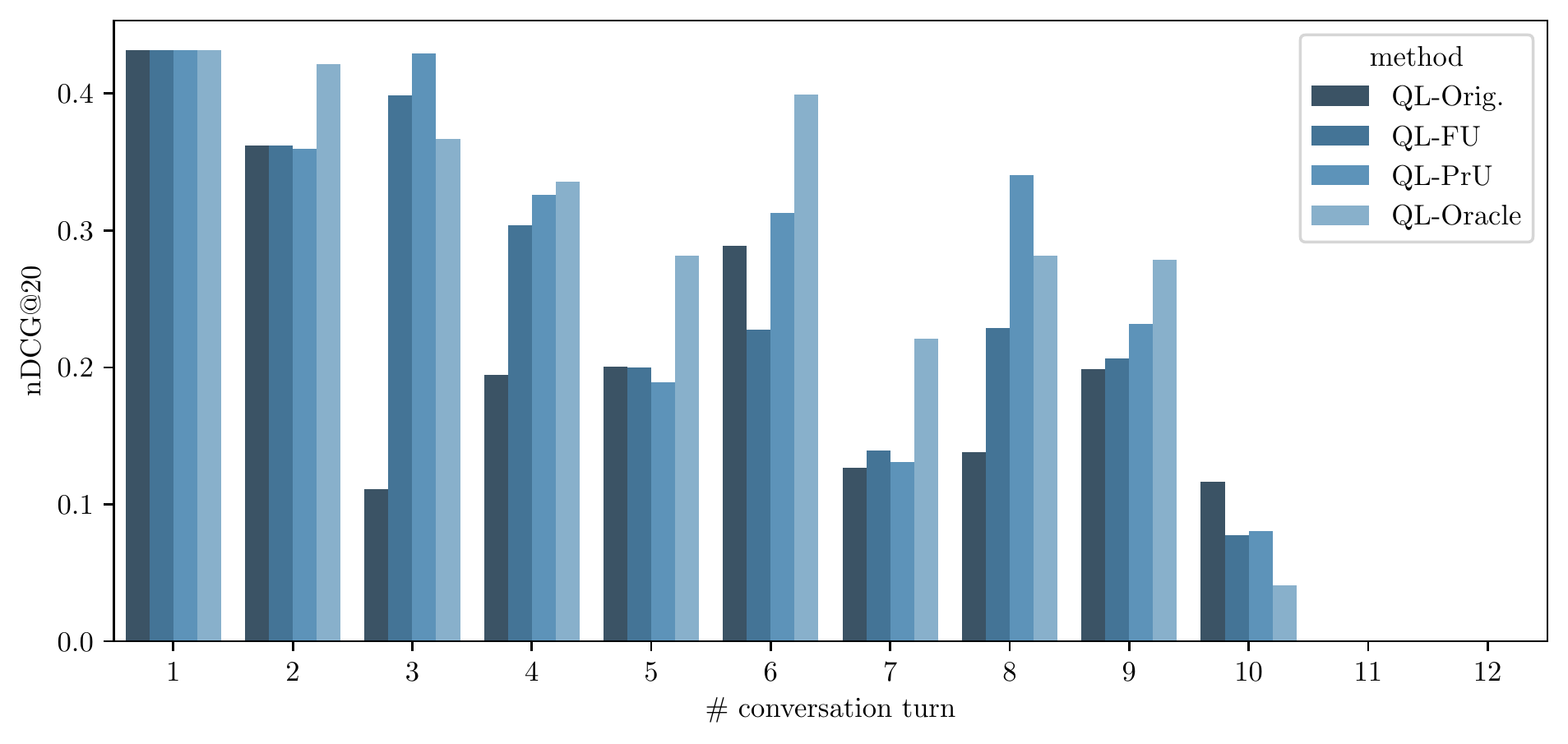}
    \vspace{-0.9cm}
    \caption{Impact of conversation turn number on the retrieval performance.}
    \label{fig:pref_turn}
    \vspace{-0.5cm}
\end{figure}

\partitle{Impact of number of turns in the conversation on the retrieval performance.} In this exercise, we study the effect of the utterance position (\ie turn) on the retrieval performance. Our aim is to understand the impact that the evolution of the conversation has during the course of a dialogue. 
Figure~\ref{fig:pref_turn} depicts the performance of QL model using different utterance reformulation strategies in terms of nDCG@20.
From Figure~\ref{fig:pref_turn}, we see that the overall performance of the system diminishes as the conversation progresses (with a few exceptions). Interestingly enough, turns 3 and 8 are positions where our proposed \modelname-based \textit{PrU} model performs better than \textit{Oracle} model. For turn 2, we observe that adding a candidate relevant question to the existing one does not really affect performance. The same is self-explanatory for the first turn. On the other hand, as the conversation reaches its end, from turn 10 onwards we observe that the potential relationship between the relevant question and the current question decreases significantly. This is in contrast with the findings of \citet{DBLP:conf/sigir/AliannejadiZCC19} where they found that as the conversation evolves the retrieval performance also increases. This is because of the different nature of the conversations in this work as the conversations do not have clarifying questions. Asking multiple clarifying questions should indeed improve the performance because it helps the system to understand the user's information need more accurately. The same does not necessarily hold in multi-turn conversations without clarifying questions because users' utterances do not necessarily elaborate more on the same information need and could diverge from the main topic of the conversation. 

\begin{figure}[t]
    \centering
    % \vspace{-0.4cm}
    \includegraphics[width=\columnwidth]{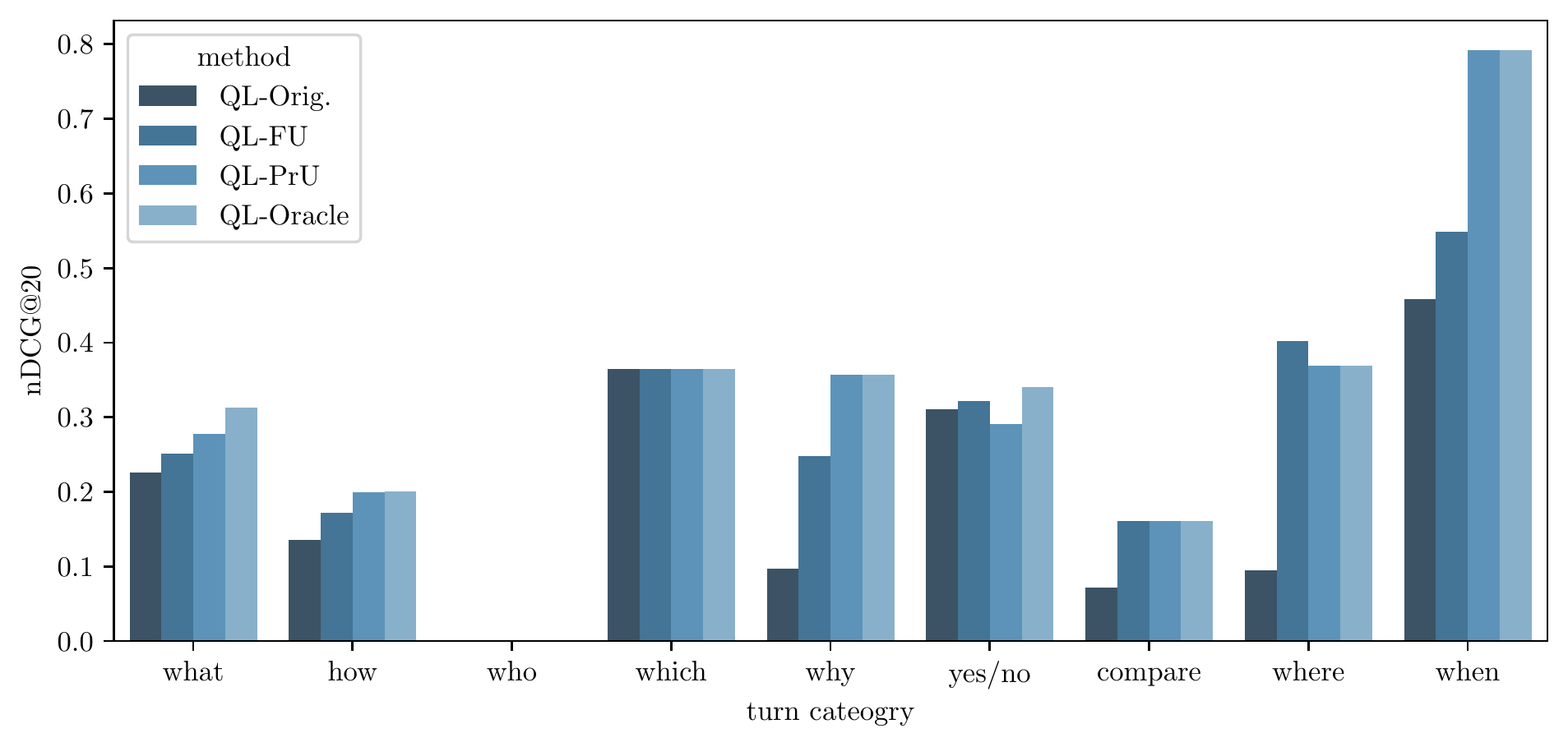}
    \vspace{-0.9cm}
    \caption{Impact of utterance (question) categories on retrieval performance.}
    \label{fig:perf_cat}
    \vspace{-0.3cm}
\end{figure}
\partitle{Impact of utterance (question) categories on retrieval performance.}
Next, we study the effect each question category has on the overall performance of the four models. Figure~\ref{fig:perf_cat} shows us that for \textit{Which} category, there is no effect of adding relevant questions to the model. This might be because there are very few utterances of \textit{Which} category and some of those utterances are conversation starters, so dependence of such questions on other questions is low. We observe for \textit{When} category remarkable improvement, at least over 60\% compared to \textit{Orig.} baseline, when either our proposed model or the oracle model is used. This can be explained from Figure \ref{fig:markov_conv}, where we see that \textit{When} category questions are always preceded by other category questions, indicating that it almost always depends on previous utterances of the conversation.

\begin{table}[!ht]
    \centering
    \vspace{-0.2cm}
    \caption{Performance comparison of passage retrieval for metric oriented relevant question selection. Best performances are marked in bold.}
    \vspace{-0.3cm}
    \begin{tabular}{lllllllll}
        \toprule
        & \multicolumn{3}{c}{\textbf{Turn Relevance}} & & \multicolumn{3}{c}{\textbf{QL based IR performance}} \\
        \cmidrule{2-4} \cmidrule{6-9}
        & \multicolumn{1}{c}{Prec.} & \multicolumn{1}{c}{Recall} & \multicolumn{1}{c}{F1} && \multicolumn{1}{c}{MAP} & \multicolumn{1}{c}{nDCG@20} & \multicolumn{1}{c}{P@20}\\
        \midrule
            PrU$_\text{P}$ & \textbf{0.79} & 0.54 & 0.64 && \textbf{0.2362} & \textbf{0.2984} & \textbf{0.1321} \\
            PrU$_\text{R}$ & 0.59 & \textbf{0.72} & 0.65 && 0.2113 & 0.1857 & 0.1108 \\
            PrU$_\text{F1}$ & 0.66 & 0.69 & \textbf{0.67} && 0.2272 & 0.2840 & 0.1229 \\
        \bottomrule
    \end{tabular}
    \label{tab:metric_pref}
    \vspace{-6mm}
\end{table}

\partitle{Impact of metric-oriented selection of relevant utterances on the retrieval.}
Our next objective is to study the effect of preference of metric (Precision vs. Recall vs. F1-measure) for relevant utterance selection on the overall performance of passage retrieval. The results of this exercise is presented in Table \ref{tab:metric_pref}. The three runs, $PrU_P$, $PrU_R$ and $PrU_{F1}$ represent the best performing run (in terms of the metric itself) for the relevance selection model for Precision, Recall and F1-measure, respectively. From the three runs, we can see that selecting relevant questions with high precision, $PrU_P$, gives us the best retrieval performance\footnote{We present only QL scorer performance since it is the best performing scorer among the three.} for all three retrieval evaluation measures. This demonstrates that for any current utterance selecting the correct relevant utterance(s) with a higher precision is preferred over selecting all the relevant utterances. We have thus recorded the performance of the precision-oriented relevant question selection based retrieval model in Table~\ref{tab:ret_res}. It is also interesting to see that $PrU_P$ achieves the best retrieval performance although it has the least recall among the three models. Also, we see that $PrU_P$ has the lowest F1-Measure, while achieving the highest retrieval performance. 

\section{Conclusions and Future Work}
In this work, we presented an extensive analysis of utterance dependency and language in multi-turn information-seeking conversations. To understand utterance dependencies we have annotated the conversations released by the TREC CAsT 2019 track. To do so, we hired three expert annotators and instructed them to identify the relevant utterance from a conversation's context. The analysis provided useful insights into the problem. For instance, we saw that there is a very strong negative correlation between relevance and distance of two utterances. This suggests that people tend to ask about information relevant to the most recent utterances. 

Furthermore, we proposed a BERT fine-tuning model to predict the relevance of the utterances. Our proposed \modelname model outperformed competitive classification baselines. Also, we demonstrated its effectiveness in improving the performance of document retrieval models. We conducted an extensive analysis on the performance of retrieval models. We found that incorporating the conversation context has positive impact on the retrieval performance. In particular, we discovered that for our proposed model the highest context dependency is found at the third turn of the conversation. Also, our experiments suggest that the task of relevant utterance prediction is highly precision-oriented, rather than being recall-oriented.
% \balance
As future work, we plan to perform a similar analysis on information-seeking conversational systems with the ability of asking clarifying questions~\cite{DBLP:conf/sigir/AliannejadiZCC19}. The proposed task in the TREC CAsT track is aiming to evaluate systems that retrieve responses to the user's utterances. However, an ideal conversational system should be able to ask questions when required. This would have a great impact on the user's behavior and definition of relevance in the conversation context. As such, we plan to extend our analysis to such conversations and investigate effective modeling of relevant utterances.

\begin{acks}
We would like to thank the three expert annotators for their invaluable help. Also, a special mention to the anonymous reviewers who helped us shape the final version of the paper. This work was partially supported by the Swiss Government Excellence Scholarships and Hasler Foundation.  %This work was partially supported by x and y projects.
\end{acks}

\bibliographystyle{ACM-Reference-Format}
% \balance
\bibliography{main}

\end{document}